\documentclass[runningheads,a4paper]{llncs}

\usepackage{amssymb}
\usepackage{graphicx}
\usepackage[utf8]{inputenc}
\usepackage{url}
\usepackage{tabularx}
\usepackage{epstopdf}
\usepackage{graphicx}
\usepackage{caption}
\usepackage{amsmath}
\usepackage{subcaption}
\usepackage{subfig} 
\usepackage{float}
\usepackage[parfill]{parskip}

\urldef{\mailsa}\path|{pplonski, zaremba}@ire.pw.edu.pl|    
\newcommand{\keywords}[1]{\par\addvspace\baselineskip
\noindent\keywordname\enspace\ignorespaces#1}

\begin{document}

\mainmatter 
\title{Image Segmentation in Liquid Argon Time Projection Chamber Detector} 
\titlerunning{Image Segmentation in LAr-TPC Detector}
\author{Piotr Płoński$^1$, Dorota Stefan$^2$, Robert Sulej$^3$, Krzysztof Zaremba$^1$}
\authorrunning{P. Płoński et al.}
\institute{$^1$Institute of Radioelectronics, Warsaw University of Technology,\\ Nowowiejska 15/19,00-665 Warsaw, Poland,\\
\mailsa
\\
$^2$Istituto Nazionale di Fisica Nucleare, Sezione di Milano e Politecnico,\\Via Celoria 16, I-20133 Milano, Italy
\\\path|dorota.stefan@ifj.edu.pl|
\\
$^3$National Center for Nuclear Research,\\A. Soltana 7, 05-400 Otwock/Swierk, Poland
\\
\path|Robert.Sulej@cern.ch| 
}
\maketitle

\begin{abstract}

The Liquid Argon Time Projection Chamber (LAr-TPC) detectors provide excellent imaging and particle identification ability for studying neutrinos. An efficient and automatic reconstruction procedures are required to exploit potential of this imaging technology. Herein, a novel method for segmentation of images from LAr-TPC detectors is presented. The proposed approach computes a feature descriptor for each pixel in the image, which characterizes amplitude distribution in pixel and its neighbourhood. The supervised classifier is employed to distinguish between pixels representing particle's track and noise. The classifier is trained and evaluated on the hand-labeled dataset. The proposed approach can be a preprocessing step for reconstructing algorithms working directly on detector images.


\end{abstract}

\section{Introduction}

The Liquid Argon Time Projection Chamber (LAr-TPC) detector idea was proposed by C.Rubbia in 1977 \cite{Rubbia}. It provides excellent imaging ability of charged particles, making it ideal for studying neutrino oscillation parameters, sterile neutrinos existence \cite{sterile}, Charge Parity violation, violation of baryonic number conservation and dark matter searches. The LAr-TPC technique is used in several projects around the world \cite{microboone}, \cite{maly_rubbia},  \cite{Argoneut}, \cite{lbne}, \cite{design-t600}. The ICARUS T600 \cite{design-t600} was the largest working detector located at Gran Sasso in the underground Italian National Laboratory operating on CNGS beam (CERN\footnote{CERN - European Organization for Nuclear Research} Neutrinos to Gran Sasso). In this study, the T600 will be used as a reference detector as other existing or planned LAr-TPC detectors have the similar construction and settings.  

In the LAr-TPC detector a charged particle produces both scintillation light and ionization electrons along its path. The scintillation light, which is poor compared to ionisation charge, is detected by photomultipliers which trigger the read-out process. Free electrons from ionizing particles drift in an uniform electric field toward the anode. They can drift to macroscopic distances because its diffussion approximate value 4.8 cm$^2$/s in highly purified liquid argon is much slower than electron drift velocity 1.59 mm/$\mu$s. The anode consists of three wire planes, so-called Induction1, Induction2, Collection. A signal is induced in a non-destructive way on the first two wire planes, which are practically transparent to the drifting electrons. The signal on the third wire plane (Collection) is formed by collecting the ionization charge. It is also the source of the calorimetric measurement. The wires in consecutive planes are oriented in three different degrees with respect to the horizontal with 3 mm spacing between wires in the plane. This allows to localize the signal source in the XZ plane, whereas Y coordinate is calcuated from wire signal timing and electron drift velocity\footnote{Coordinate system labelling is given for reference.}. Signal on wires is amplified and digitized with 2.5 MHz sampling frequency which results in 0.64 mm spatial resolution along the drift coordinate. The digitized waveforms from consecutive wires placed next to each other form a 2D projection images of an event. The image size resolution is 0.64 mm x 3 mm.

The registered images are input for reconstruction and particle identification algorithms, which estimate physical quantities from events observed in the detector. To our knowledge, all recently presented analysis assume that images are preprocessed into so-called set of hits \cite{Argoneut}, \cite{warwick}, \cite{3d-track}. Each hit represents a position of signal on wire coming from particle's track. Hit has information about the beginning, the end and the peak of the particle's track signal on wire. Hits are obtained by fitting a multiple reference signal pulses on wire's signal. This approach has some limitations. The hit fitting algorithm distinguishes signal from particle's track and noise based on defined threshold value, which can fail in case of noise with high amplitude. This can be a common case during analysis of images deconvoluted with impulse response of the wire signal readout chain, because the signal-to-noise ratio descends. What is more, the hit unit can be imprecise in case of multiple tracks overlapping in the same position in the image. These drawbacks can be overcame by constructing algorithms working directly on raw images. In this paper, the novel method for images segmentation from LAr-TPC detectors is presented, which can be used as a preprocessing step for reconstruction and identification algorithms working directly on images from detector. Furthermore, this segmentation method can be used to improve hit fitting procedure by replacing identification of particle's track based on threshold value. The presented approach use a supervised classifier working on a feature descriptor which characterizes the amplitude distribution in pixel and its neighbourhood. The classifier is trained and evaluated on a hand-labeled dataset. The performance of the proposed method is compared with segmentation based on threshold value. The importance of proposed features is studied. 

\section{Methods}

\subsection{Proposed Feature Descriptor} \label{features}

Describing image by set of features is a common approach in computer vision \cite{lowe}, \cite{rafal1}, \cite{rafal2}, \cite{rafal3}, \cite{rafal4}. The feature vector is constructed to obtain the relevant information from the image data. Herein, in order to characterize distribution of signal in the pixel and its neighbourhood a feature descriptor is constructed, consisting of 42 features, which are described below:
\begin{itemize}
\item Pixel value, proportional to the charge value;
\item Signal statistics, which describe the amplitude distribution in the pixel and its neighbourhood, represented by: maximum, minimum, median, mean and standard deviation computed for 3x3, 5x5 and 7x7 kernel sizes;
\item Difference of Gaussians, which is substraction of two blurred images with Gaussian kernel with different standard deviations, which can be viewed as band-pass filter. Used pairs of standard deviations are : $\{0.5, 2\}$, $\{0.5, 3\}$, $\{0.5, 4\}$, $\{0.75, 2\}$, $\{0.75, 3\}$, $\{0.75, 4\}$, $\{1, 2\}$, $\{1, 3\}$, $\{1, 4\}$; 
\item Image gradient magnitude, which detects sudden changes in the amplitude, the Prewitt operator was used for gradient computation;
\item Ordered eigenvalues of image's Hessian as well as sum and multiplication of them, which describes the second order of local image intesity;
\item Ordered eigenvalues of image's tensor, together with sum and multiplication of them for kernel sizes 3x3, 5x5, 7x7, which indicates the strength of the directional intensity change. 
\end{itemize}
The similar configuration of proposed features is often used in image analysis in medicine \cite{sbfsem}, \cite{rudzki}. Recently, the usage of  image tensor eigenvalues was proposed for detecting points of interest and delta electrons in images from LAr-TPC detector \cite{points_ben}.

\subsection{Datasets creation}

%icat600_nwq_2000mev001
To generate datasets the FLUKA software \cite{fluka} and T600 detector parameters were used. The events were generated for 2 GeV energy. From obtained events 50 images for Induction2 and Collection views were independently selected. The two views were chosen to present the performance of the proposed method on two sources of images which have different characteristics of the signal. All images were deconvoluted with impulse response of wire signal readout chain of each plane and resized to have similar resolution on both axis. The dedicated application with Graphical User Interface was used to manually label the dataset. The pixels representing particle's track are noted as a positive class whereas noise pixels as a negative class. For each pixel in the dataset the feature vector was computed according to the description in Section \ref{features}. The feature vector together with class label of the pixel form a sample in the dataset. The dataset was divided into train and test subsets. There are 40 events in the train subset and 10 events in the test subset in each considered plane. The number of samples in each subset for Induction2 and Collection views are presented in Table \ref{datasets}. They are different because the readout procedure is triggered by photomultipliers only on subset of wires, from which particle's track is registered. The classes are highly imbalanced in considered datasets. The ratios of positives to negatives in train dataset are 1:305 and 1:109 for Induction2 and Collection views, respectively. 

\begin{table}[ht]
\vspace*{-4mm}
\def\arraystretch{1.25}
\centering
\begin{tabular}{rrrrr}
  \hline
 & \multicolumn{2}{c}{Induction2}  & \multicolumn{2}{c}{Collection} \\
 \hline
 & \ \ Negative \ \ & \ \ Positive  & \ \ Negative \ \ & \ \ Positive \\
 Train & 15,648,122 & 51,300 & 7,290,844 & 66,749 \\
 Test & 4,427,377 & 10,437 & 2,061,778 & 16,988 \\
  \hline
\end{tabular}
\vspace*{4mm}
\caption{The number of samples in train and test subsets for Induction2 and Collection views. Positive samples represent particle's track pixels whereas negative samples represent noise.} \label{datasets}
\end{table}

\subsection{Classifiers}

In the proposed approach, the created feature descriptor is an input for the classifier, which response is a probability whether pixel represents particle's track in the image. The Logistic Regression (LR) \cite{elements} and Random Forest \cite{breiman}, \cite{rf-som} algorithms were used as classifiers. They were trained on all features described in Section \ref{features} contrary to the Decision Stump (DS) \cite{stump} trained only on considered pixel's amplitude value. The performance of the DS is baseline because it is commonly used in analysis by physicists. To asses the classifier's performance the precision recall graph was used, where 
\begin{equation}
Precision = \frac{TP}{TP+FP},  
\end{equation}
\begin{equation}
Recall = \frac{TP}{TP+FN}.
\end{equation}
The TP stands for true positives - correctly classified positive samples, FP are false positives - negative samples incorrectly classified, and FN are false negatives, which are positive samples improperly classified as negatives.

\begin{figure}[H]
\captionsetup[subfigure]{aboveskip=-13pt,belowskip=-3pt}
\centering
\begin{subfigure}[b]{0.47\textwidth}
	\caption{Decision Stump, Induction2}  
	\includegraphics[clip=true, width=\textwidth]{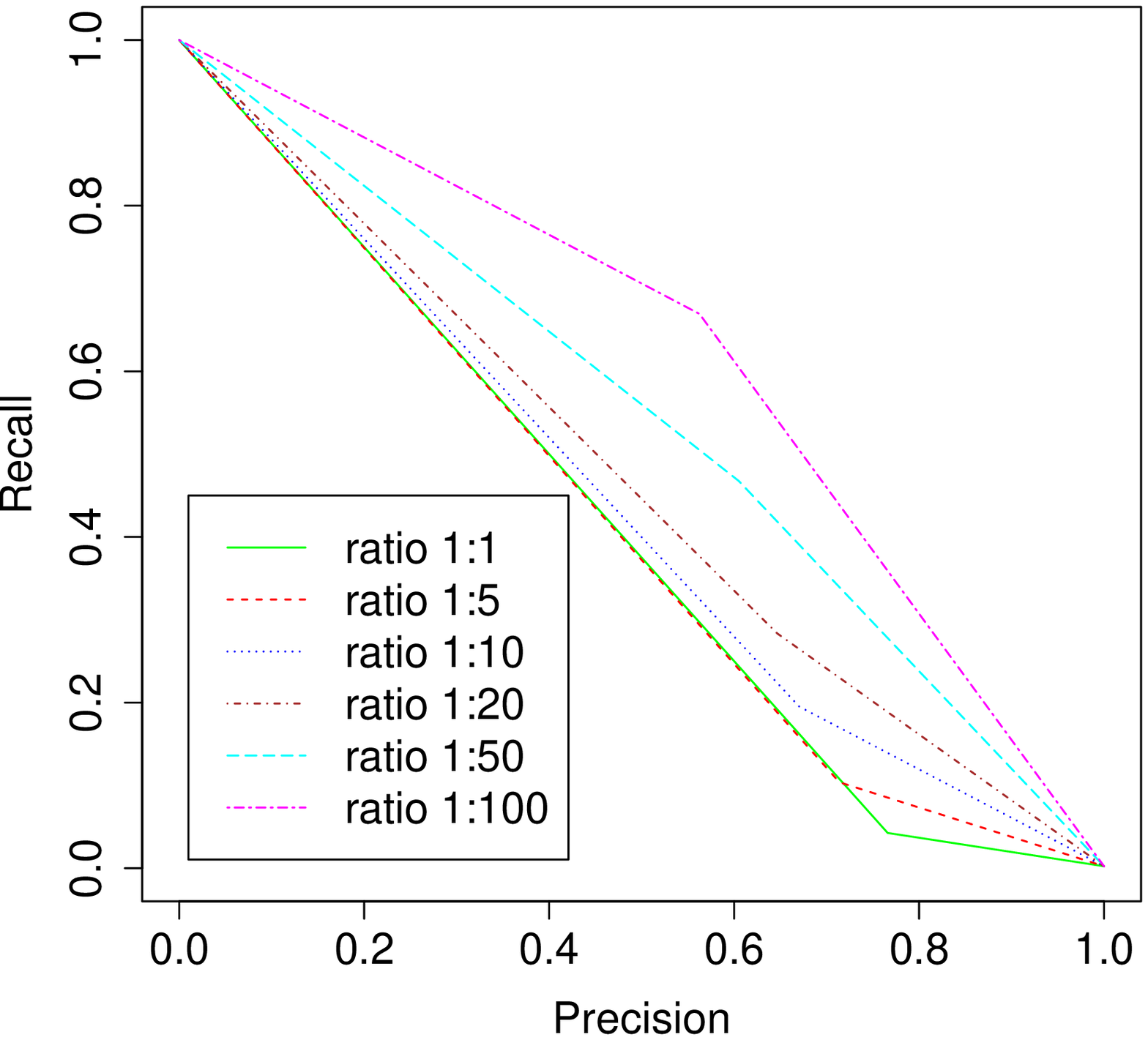}\end{subfigure}
\begin{subfigure}[b]{0.47\textwidth}
	\caption{Decision Stump, Collection} 
	\includegraphics[clip=true, width=\textwidth]{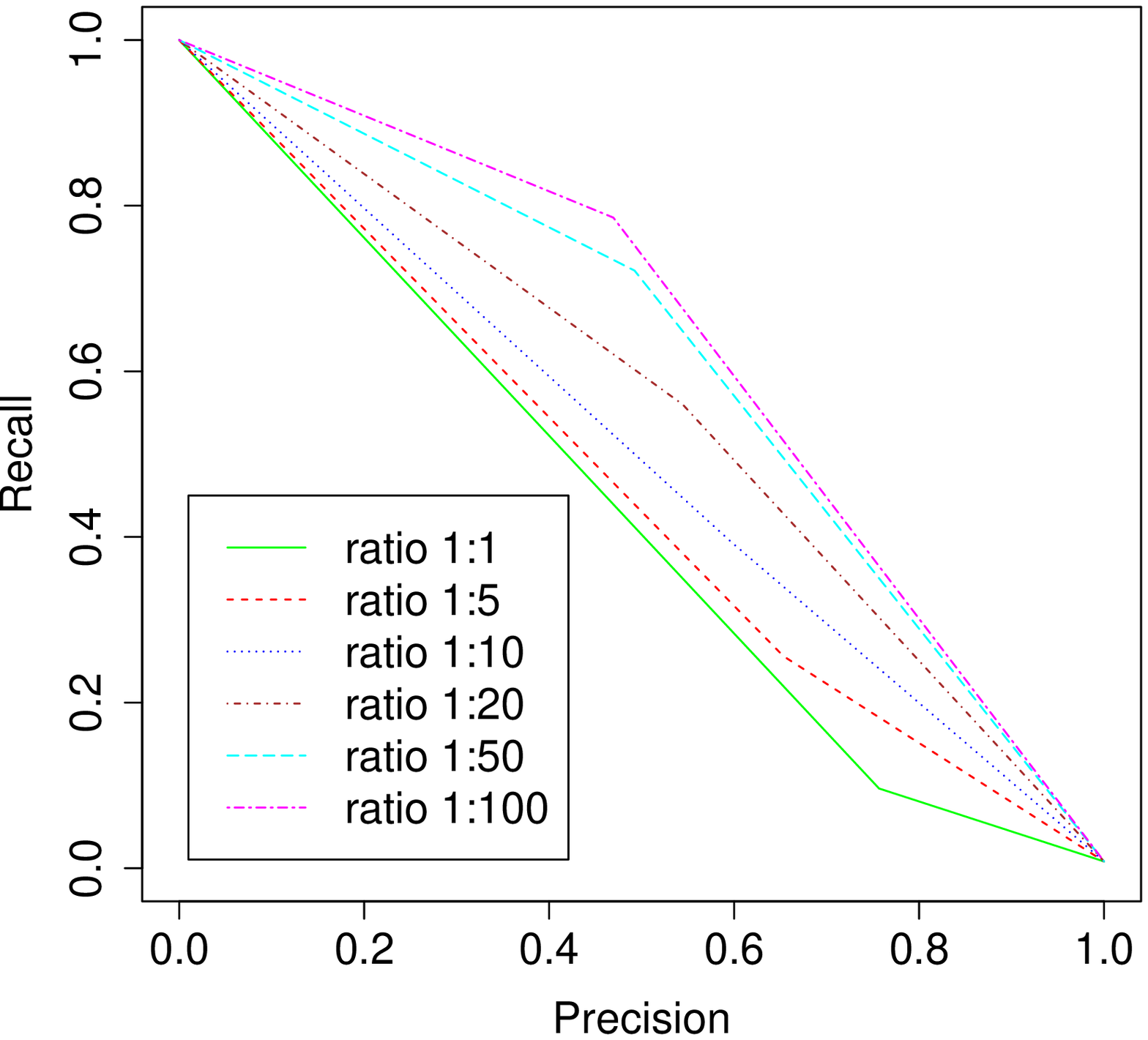}
\end{subfigure}
\begin{subfigure}[b]{0.47\textwidth}
	\caption{Logistic Regression, Induction2} \label{mds_pi}
	\includegraphics[clip=true, width=\textwidth]{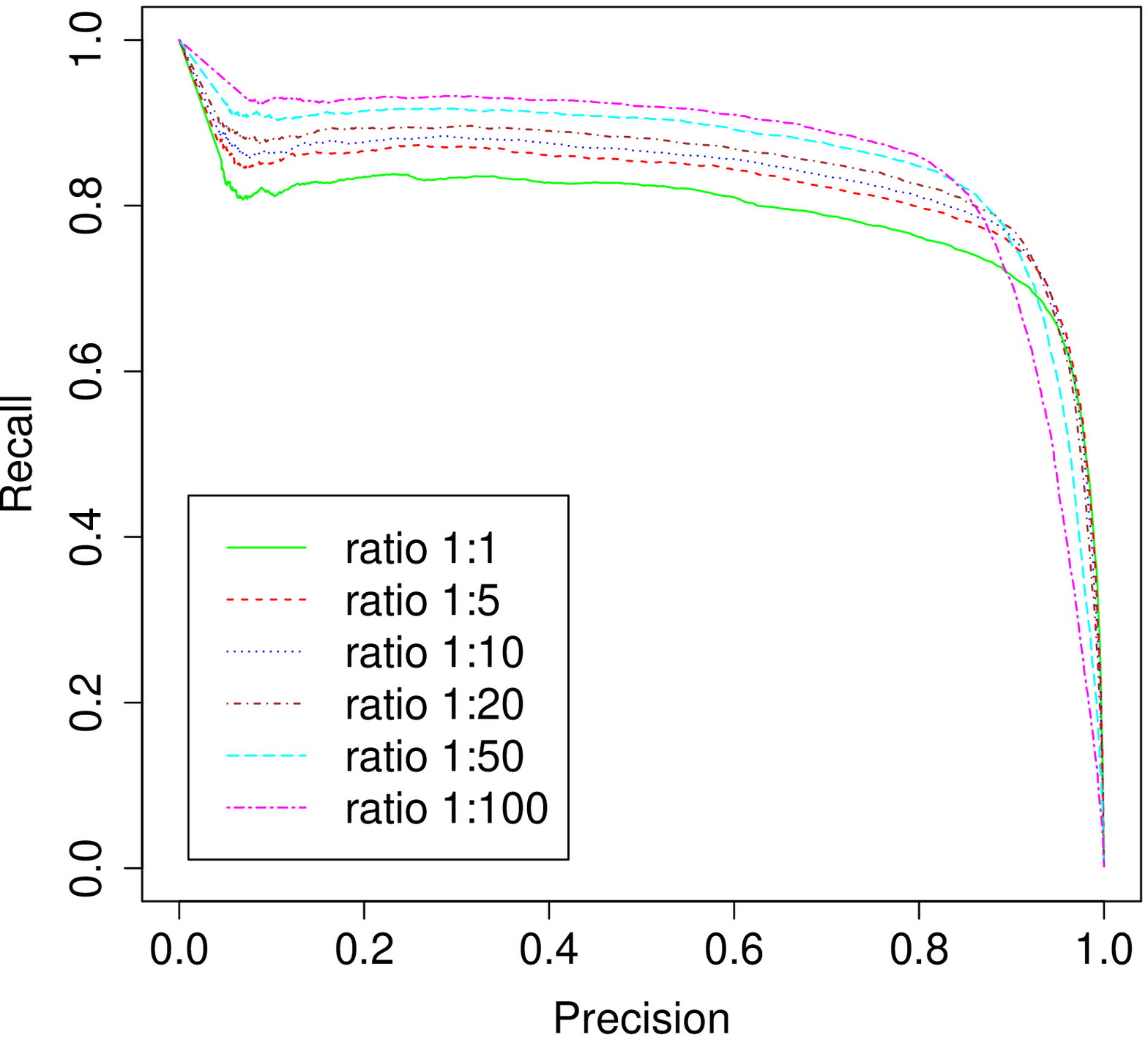}\end{subfigure}
\begin{subfigure}[b]{0.47\textwidth}
	\caption{Logistic Regression, Collection} 
	\includegraphics[clip=true, width=\textwidth]{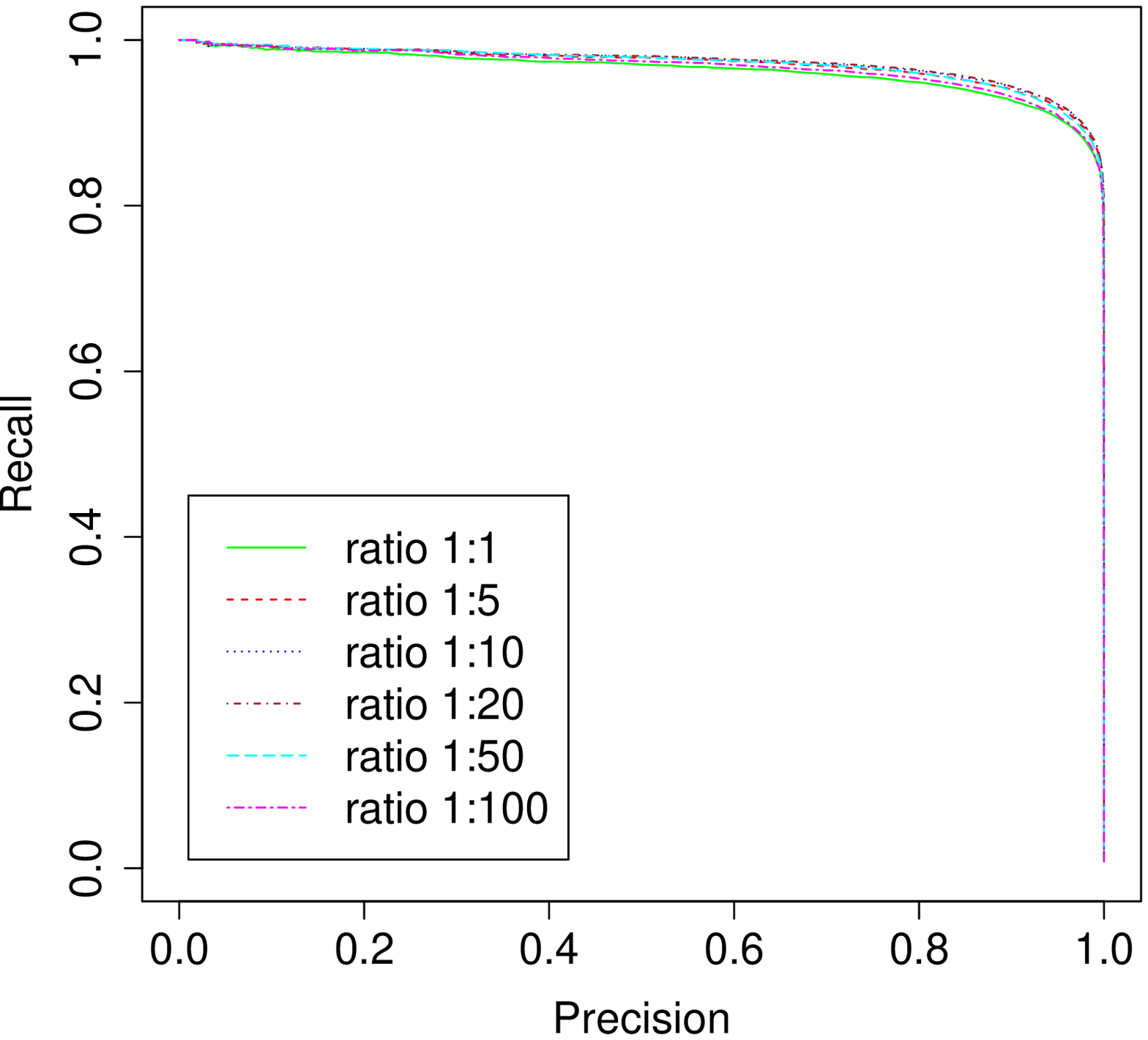}
\end{subfigure}
\begin{subfigure}[b]{0.47\textwidth}
	\caption{Random Forest, Induction2}  
	\includegraphics[clip=true, width=\textwidth]{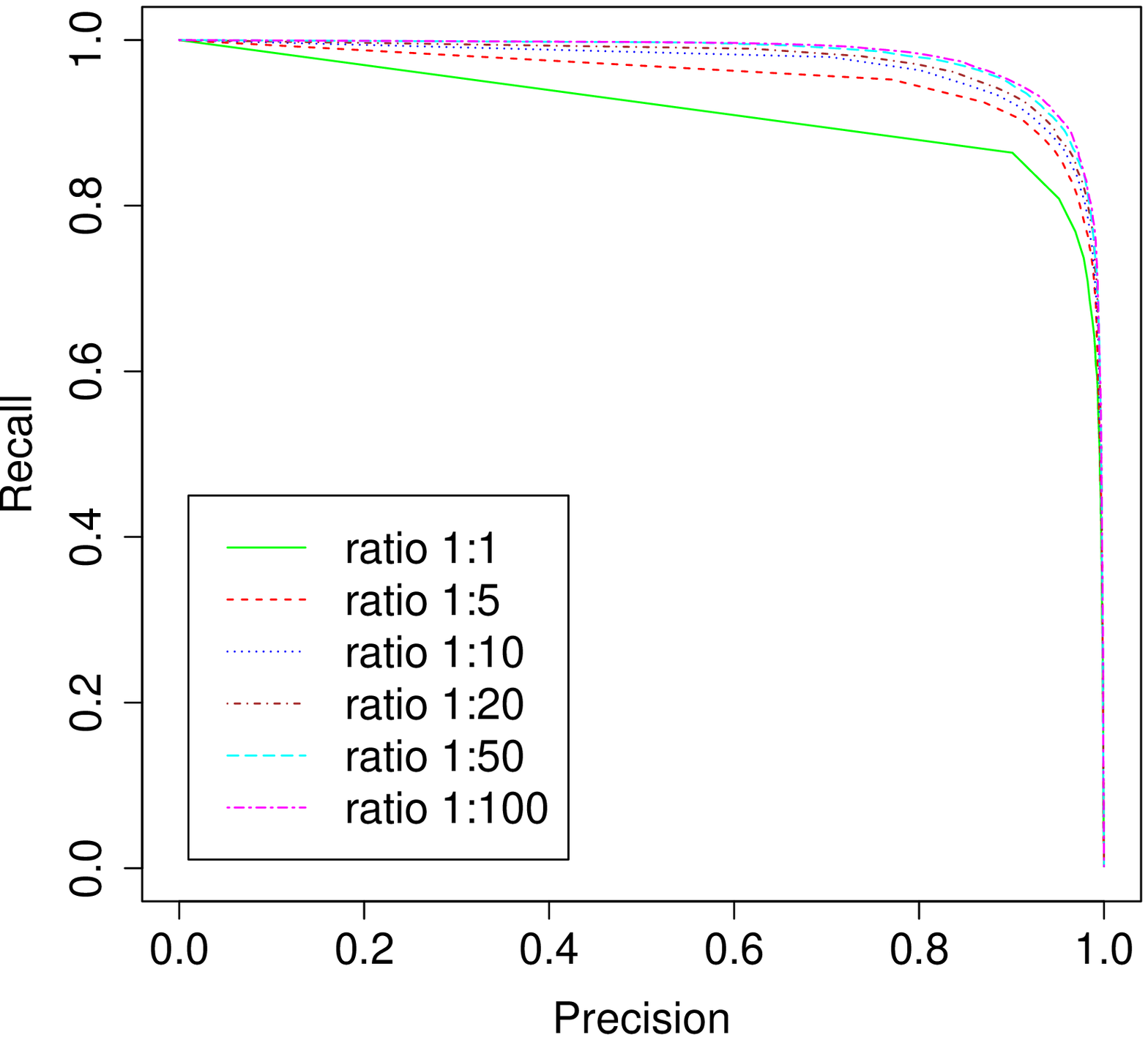}\end{subfigure}
\begin{subfigure}[b]{0.47\textwidth}
	\caption{Random Forest, Collection}  
	\includegraphics[clip=true, width=\textwidth]{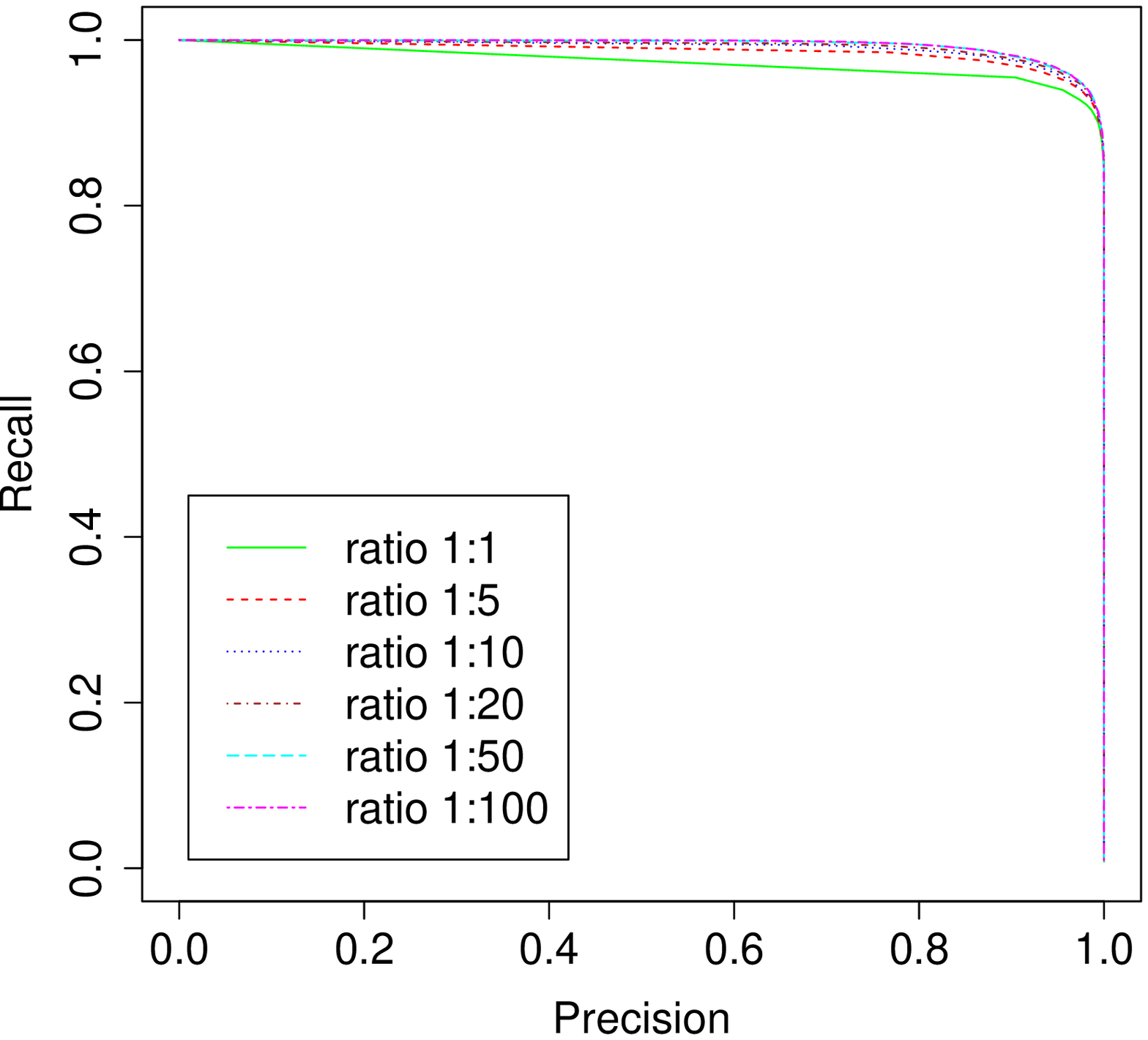}
\end{subfigure}
\caption{The precision recall graphs for Decision Stump, Logistic Regression and Random Forest classifiers for Induction2 and Collection views. Classifiers were trained with different ratios of positive to negative classes in train set and evaluated on all samples from test set.} \label{class_ratios}
\end{figure}

\section{Results}
 
\begin{figure}[H]
\captionsetup[subfigure]{aboveskip=-11pt,belowskip=-13pt}
\centering
\begin{subfigure}[b]{0.49\textwidth}
	\caption{Induction2}  
	\includegraphics[clip=true, width=\textwidth]{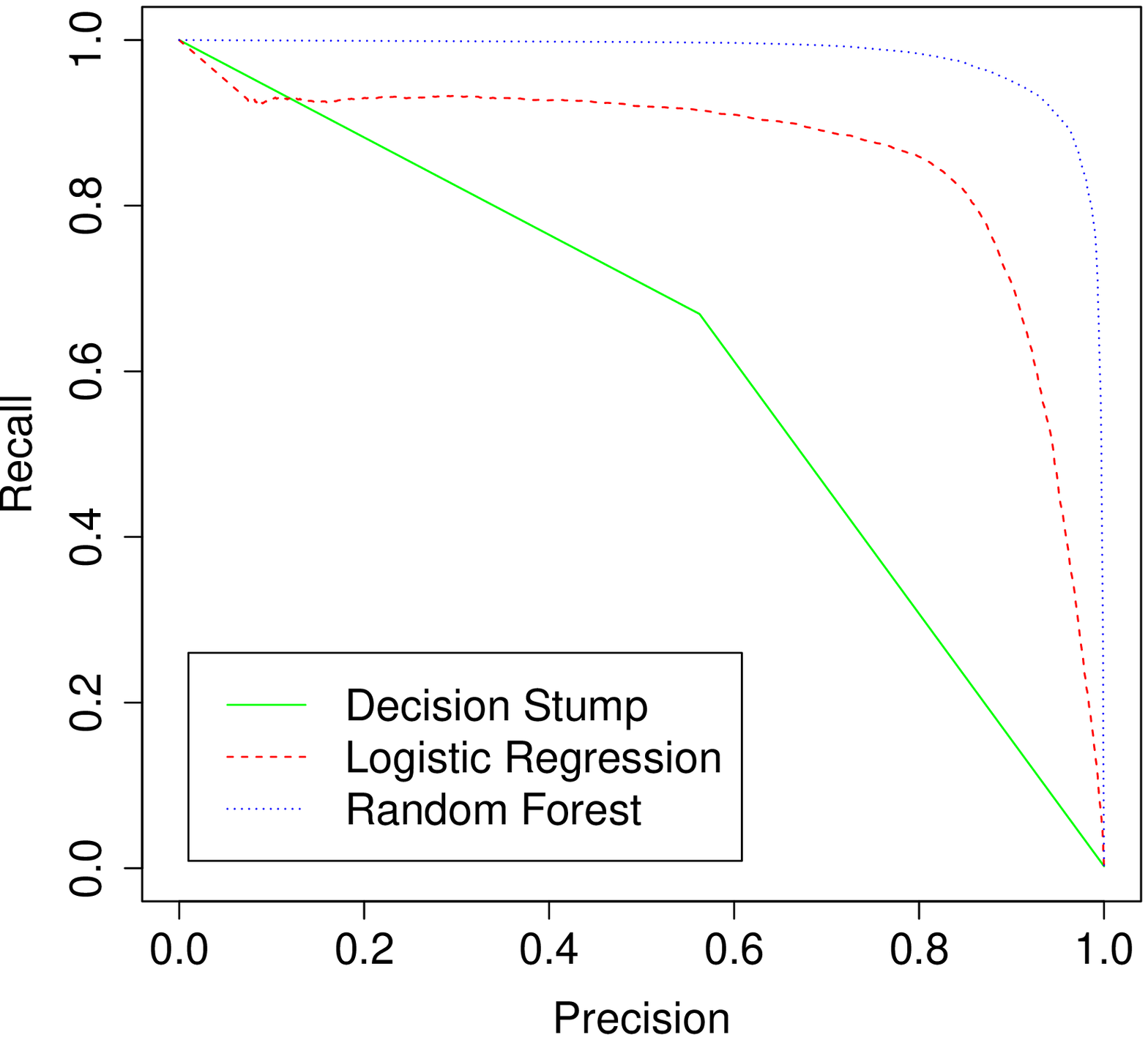}\end{subfigure}
\begin{subfigure}[b]{0.49\textwidth}
	\caption{Collection}  
	\includegraphics[clip=true, width=\textwidth]{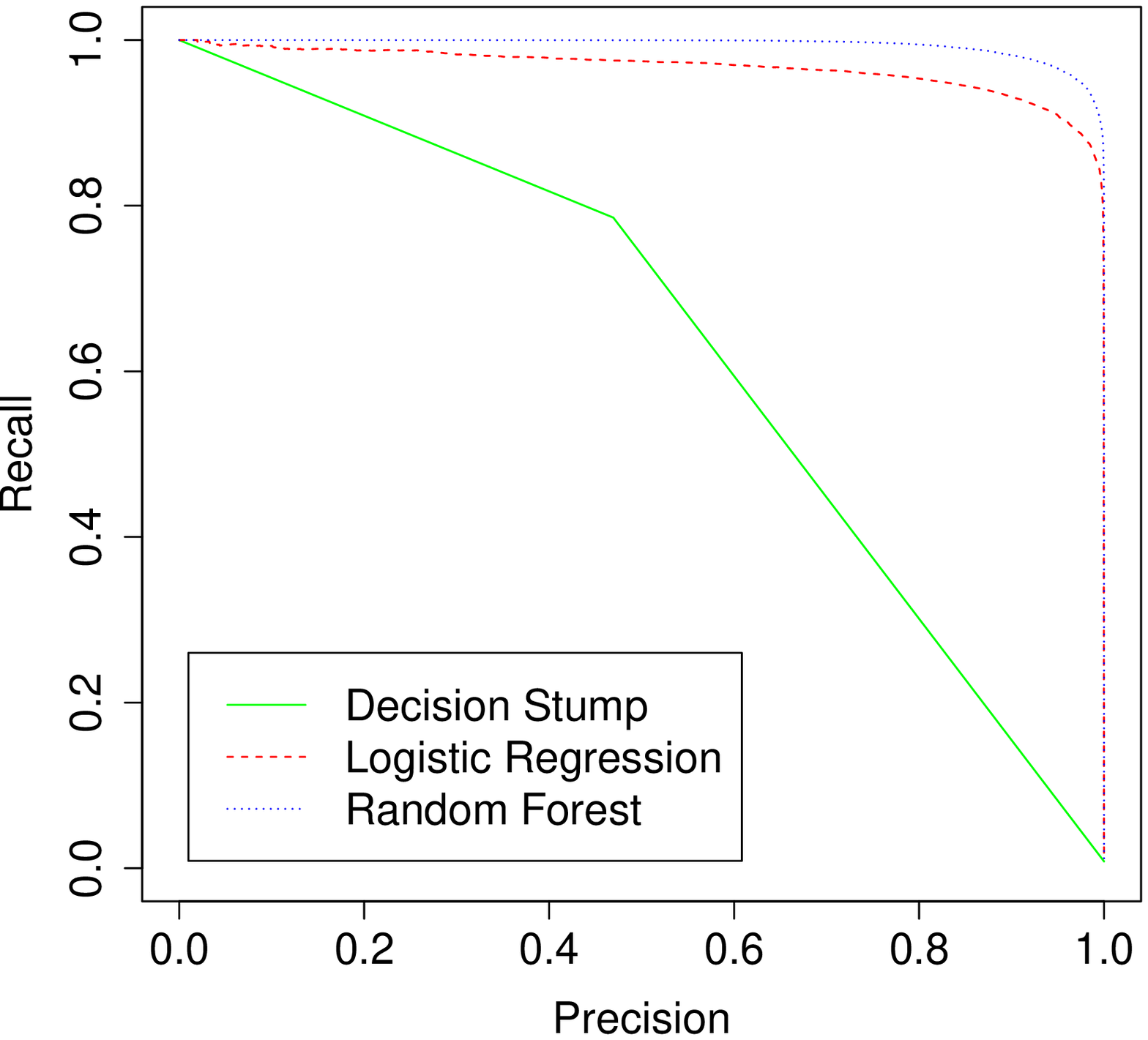}
\end{subfigure}
\caption{The comparison of classifiers preformance learned with train set with class ratio 1:100 of positive to negative samples and evaluated on all test samples, for Induction2 and Collection views.} \label{comp_roc}
\end{figure}

Due to the facts that classes are highly imbalanced and there is a large number of samples, the train datasets were downsampled. The classifiers were trained on different class ratios: 1:1, 1:10, 1:20, 1:50, 1:100 - with constant number of positive samples and changing the number of negative samples. All the samples from the test set were used in evaluation of the classifiers performance. There were used 100 trees in RF algorithm for both views, later the selection of optimal number of trees in the forest is reported. The precision recall graphs for DS, LR and RF are presented in the Fig.\ref{class_ratios}. The accuracy of DS classifier increases with increasing number of negative samples in the train set because the threshold value in decision rule can be more precisely estimated for larger dataset. For LR and RF the similar behavior can be observed, especially for Induction2. The more data is available the more accurate classifiers can be trained. From this point, the 1:100 classes ratio is used in tha analysis for both views. The comparison of used classifiers' performance is presented in the Fig.\ref{comp_roc}. The RF classifier outperforms DS and LR on both views. The RF algorithm, which is an ensemble of unpruned decision trees, is able to learn more about data than DS and LR. The difference in performance between RF and other algorithms is larger on Induction2 than on the Collection. Therefore, the Induction2 signal seems to be more complex to classify than Collection signal.

\begin{figure}[H]
\captionsetup[subfigure]{aboveskip=-11pt,belowskip=-13pt}
\centering
\begin{subfigure}[b]{0.49\textwidth}
	\caption{Induction2} \label{mds_pi}
	\includegraphics[clip=true, width=\textwidth]{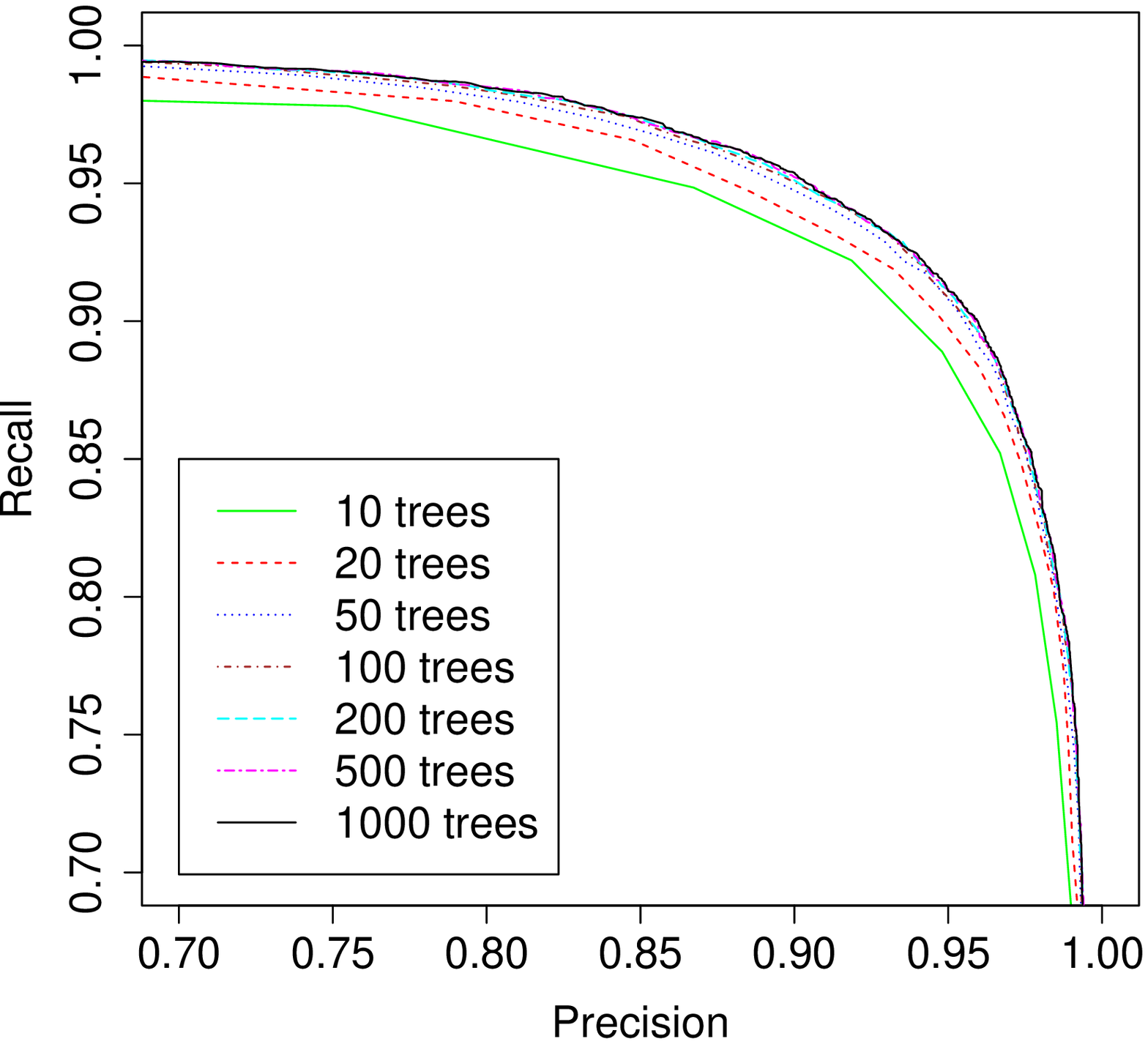}\end{subfigure}
\begin{subfigure}[b]{0.49\textwidth}
	\caption{Collection} \label{rfmds_pi}
	\includegraphics[clip=true, width=\textwidth]{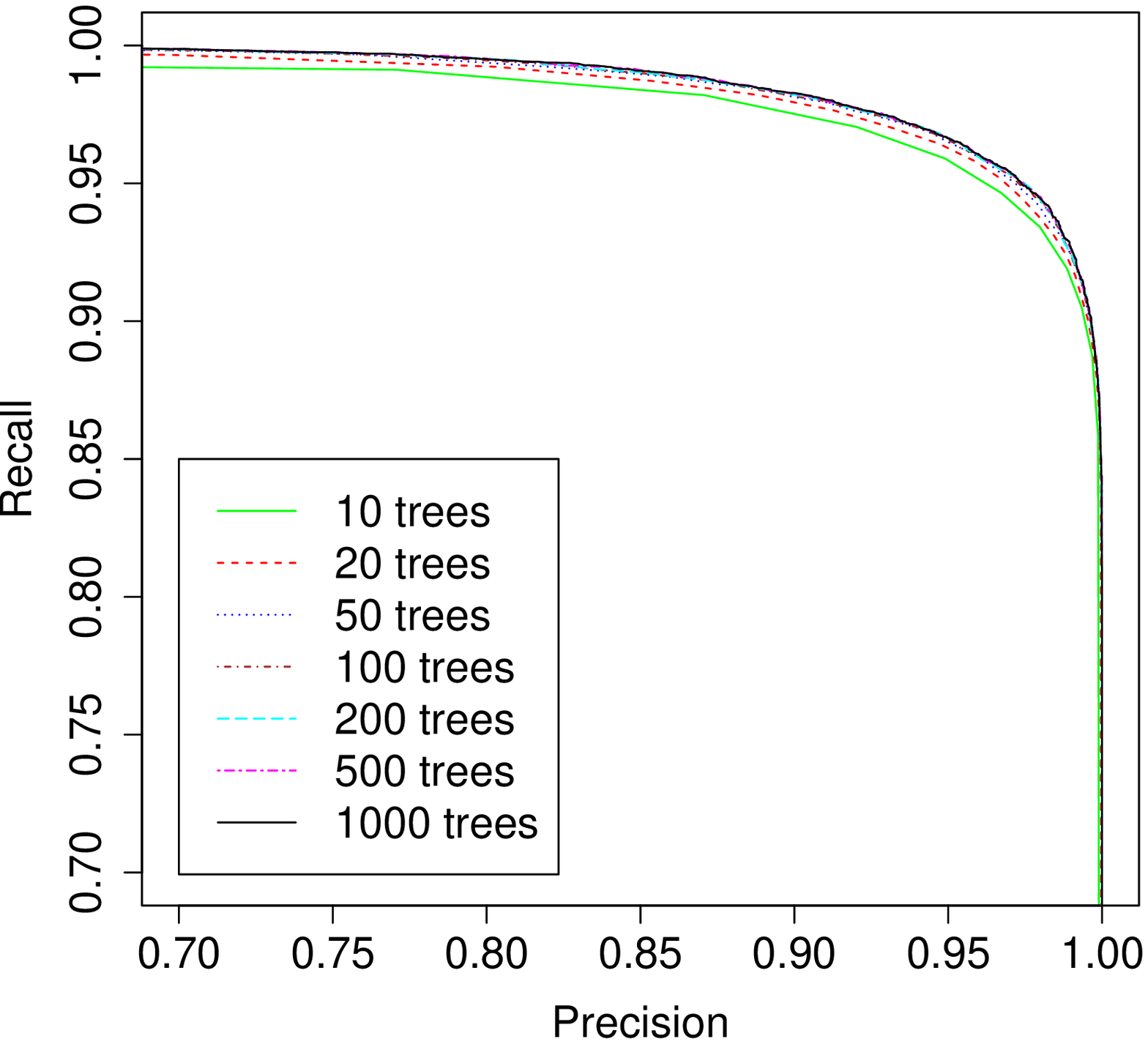}
\end{subfigure}
\caption{The precision recall graphs for Random Forest algorithm for different number of trees used, trained on class ratio of positives to negatives 1:100 and evaluated on all test samples for Induction2 and Collection views. } \label{rf_trees}
\end{figure}

The RF performance for different number of trees was examined. The precision recall graphs for trees number $\{10, 20, 50, 100, 200, 500, 1000\}$ for class ratio 1:100 for Induction2 and Collection are presented in Fig.\ref{rf_trees}. The RF achieves the highest performance starting from 100 trees in the ensemble for Induction2 and the classifier accuracy does not significantly change with adding more trees. Whereas, the classification performance is the highest starting from 50 trees in the forest for Collection view. The higher number of trees in the forest is needed to learn a dataset in the Induction2 than in the Collection view. This once more indicates that Induction2 signal is more complex in classification than the Collection signal.

\begin{figure}[H]
\vspace*{-6mm}
\captionsetup[subfigure]{aboveskip=1pt,belowskip=1pt}
\centering
\begin{subfigure}[b]{0.4\textwidth}
	\caption{Raw signal} 
	\fbox{\includegraphics[clip=true, width=\textwidth]{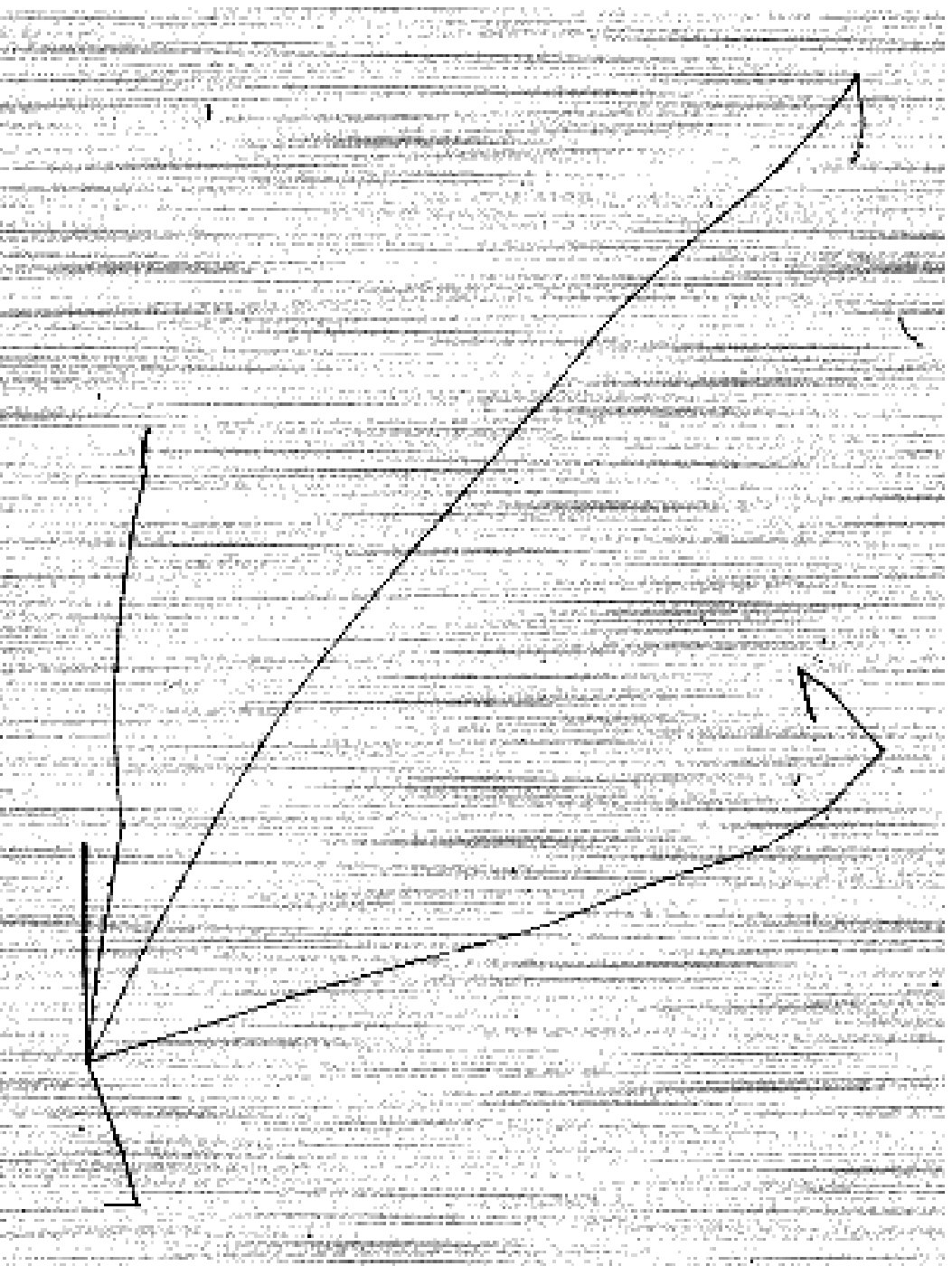}}
\end{subfigure}
\begin{subfigure}[b]{0.4\textwidth}
	\caption{Decision stump}
	\fbox{\includegraphics[clip=true, width=\textwidth]{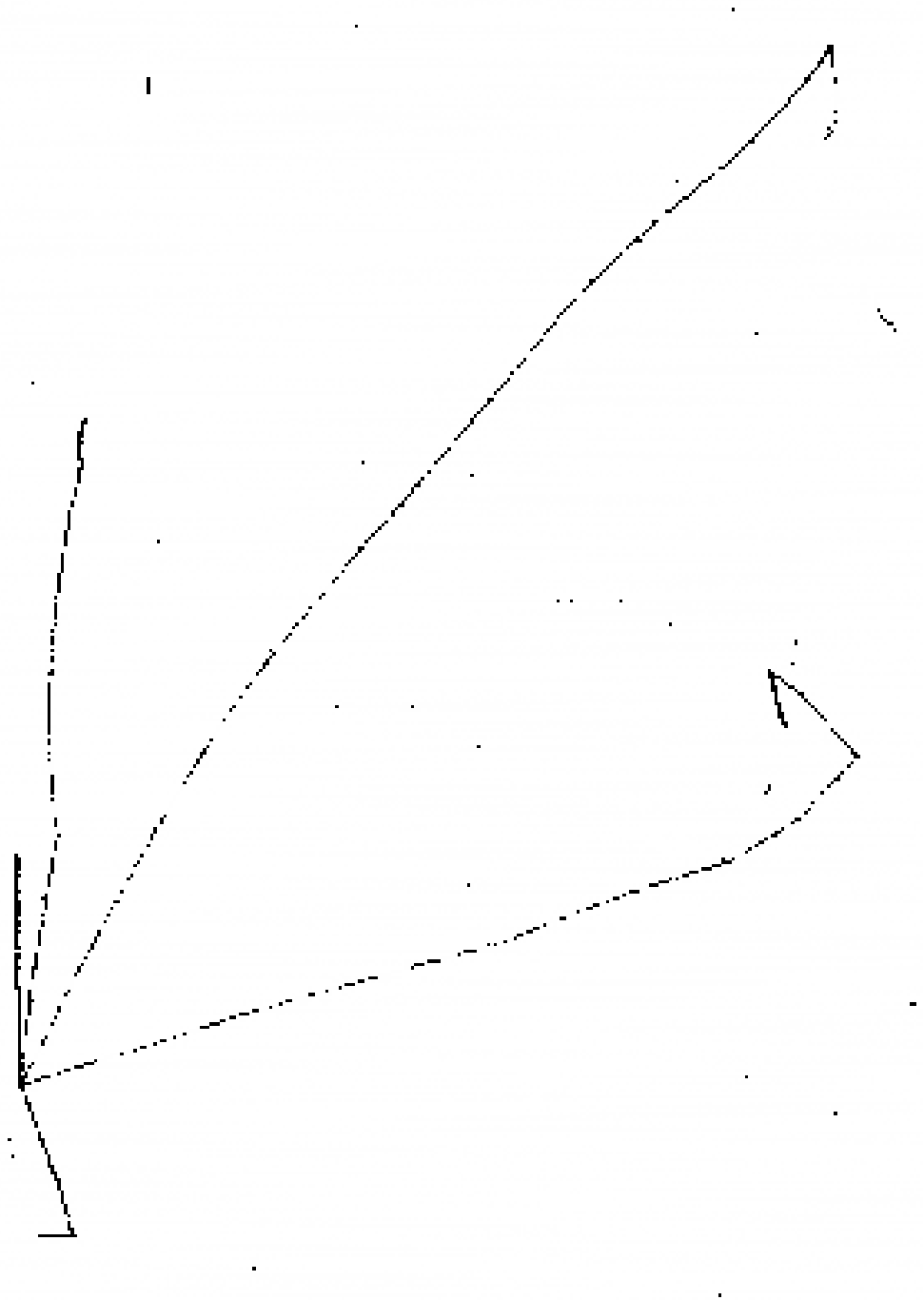}}
\end{subfigure}
\begin{subfigure}[b]{0.4\textwidth}
	\caption{Logistic Regression} 
	\fbox{\includegraphics[clip=true, width=\textwidth]{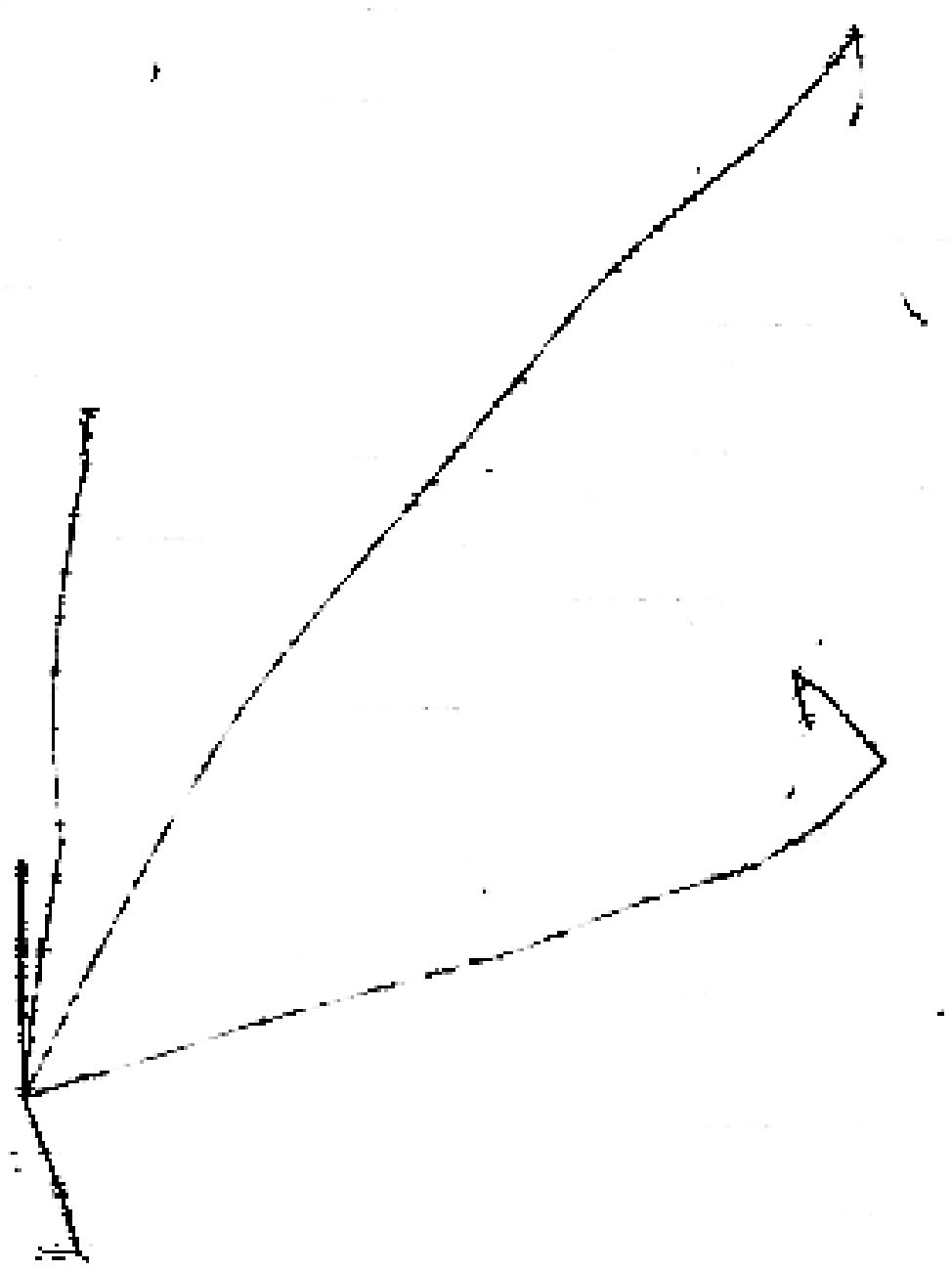}}
\end{subfigure}
\begin{subfigure}[b]{0.4\textwidth}
	\caption{Random Forest}
	\fbox{\includegraphics[clip=true, width=\textwidth]{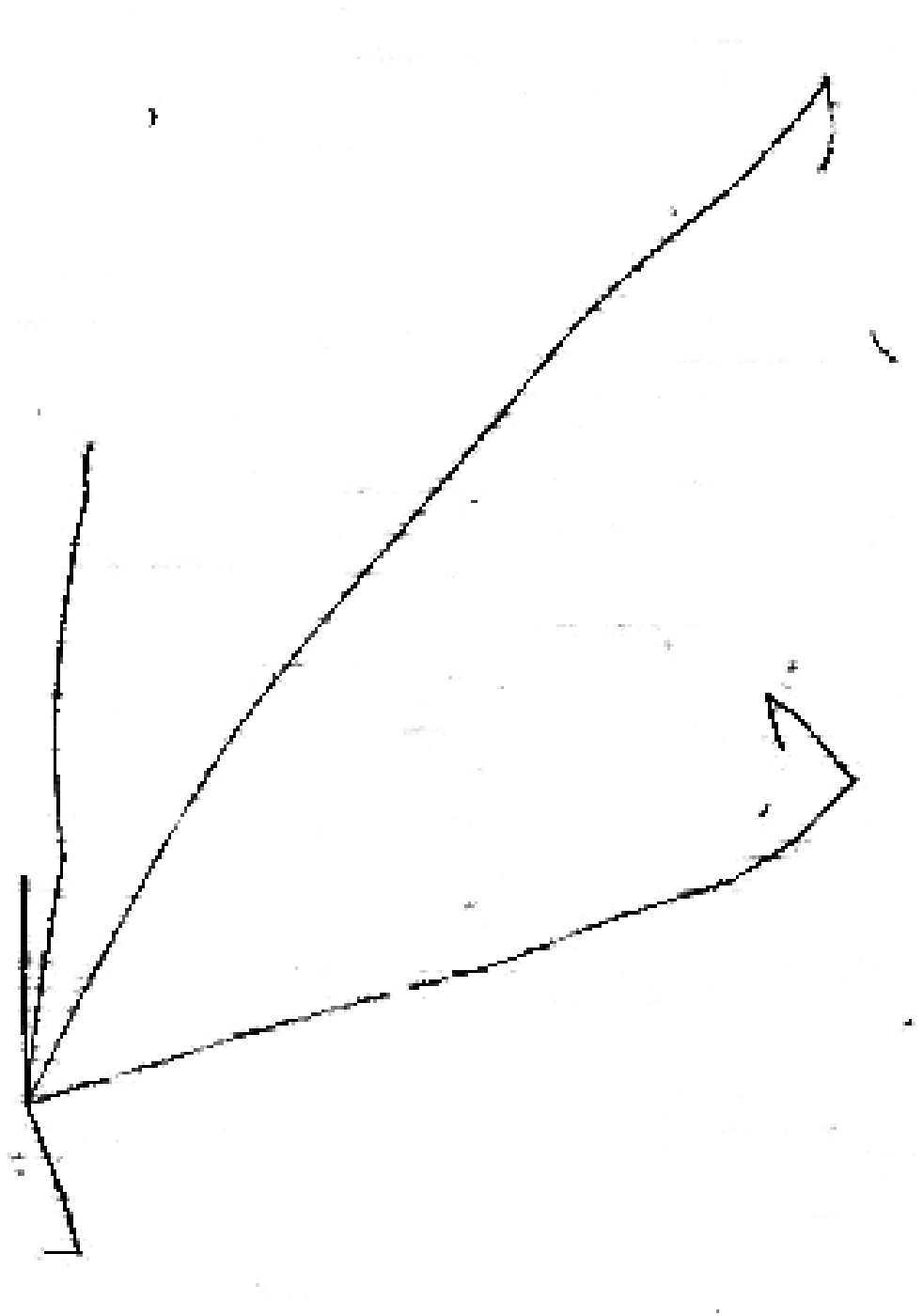}}
\end{subfigure}
\caption{(a) The raw signal of test event from Induction2.  (b,c,d) The response of the classifiers for considered test event.  The classifiers were trained on classes ratio 1:100.}\label{response}
\end{figure}

The response of the classifier is a probability that pixel represents particle's track. The examples of all examined classifiers responses on the test event from Induction2 are presented in the Fig.\ref{response}. It can be observed that for DS and LR there are more breaks on tracks than on response from RF. Therefore, the RF output is preferable for further reconstructing algorithms. 

One of the key features of the RF algorithm is that the variable importance can be easily obtained from learned model \cite{breiman}. The top ten features for Induction2 and Collection views are listed in Table \ref{var_imp}. It is worth to notice, that there are different features selected as the most important for Induction2 and Collection. It is interesting that pixel amplitude was selected as the ninth of the most important features for Induction2 and is not present in the top ten important features for Collection dataset. Although based on only this variable the decision of pixel representation was done in the threshold method. All the proposed features described in Section \ref{features} used in the RF training are important in classification, since elimination of the least important features decreases the classifier performance on each view.

\begin{table}[ht]
\centering
\begin{tabular}{rrr}
  \hline
\ \ \  Rank & \ \ \ Induction2 \ \ \ & \ \ \ Collection \ \ \  \\ \hline
 1 & maximum in 5x5 kernel & mean in 3x3 kernel \\
 2 & maximum in 3x3 kernel & median in 3x3 kernel \\
 3 & maximum in 7x7 kernel & mean in 5x5 kernel \\
 4 & standard deviation in 5x5 kernel & tensor 1$^{st}$  eigenvalue in 3x3 kernel\\
 5 & standard deviation in 3x3 kernel & median in 5x5 kernel \\
 6 & tensor 1$^{st}$ eigenvalue in 3x3 kernel & maximum in 5x5 kernel \\
 7 & standard deviation in 7x7 kernel & standard deviation in 5x5 kernel \\
 8 & mean in 3x3 kernel & mean in 7x7 kernel \\
 9 & pixel amplitude & maximum in 3x3 kernel \\
 10 & median in 3x3 kernel & maximum in 7x7 kernel \\
 \hline
\end{tabular}
\caption{The top ten important features according to the Random Forest algorithm with 100 trees trained on classes ratio 1:100 for Induction2 and Collection views.} \label{var_imp}
\end{table}

\section{Conclusions}

The efficient computerized methods for automatic analysis of observed events are required to fully exploit the imaging potential of LAr-TPC detectors. Herein, the novel method for image segmentation is presented. In the proposed approach the feature descriptor for each pixel in image is computed. It describes the distribution of signal in the pixel and its neighbourhood. Based on constructed features the classifier makes a decision whether pixel represents particle's track or noise. The two popular classifiers were examined, namely: Logistic Regression and Random Forest. The classifier was trained and evaluated on the hand-labeled dataset on images from two distinct views. The proposed method outperforms with large margin the widely used method of image thresholding based on pixel amplitude on both Induction2 and Collection views. The method is an universal and can be used for various characteristic of the signal in the image after the classifier training. The proposed method can be used as a preprocessing step for reconstructing algorithms working directly on raw images. On the next step of event analysis pixels in close neighbourhood with positive class label will be clustered into tracks. The performance of the method can be further improved by using other classifiers algorithms, like Multi-Layer Perceptron \cite{mlp} or extending the feature vector. It would be interesting to compare the proposed method working on prepared feature descriptor with method working on automatically constructed features from Convolutional Neural Network \cite{jurgen}.

\section{Acknowledgements}
PP and KZ acknowledge the support of the National Science Center (Harmonia 2012/04/M/ST2/00775). Authors are grateful to the ICARUS Collaboration and Polish Neutrino Group for useful suggestions and constructive discussions during a preliminary part of this work.

\end{document}